\documentclass{article}
\usepackage{ijcai19}

\usepackage{times}
\usepackage{soul}
\usepackage{url}
\usepackage[small]{caption}
\usepackage{graphicx}
\usepackage{footmisc}
\usepackage{amsmath}
\usepackage{booktabs}
\usepackage{algorithm}
\usepackage{algorithmic}
\usepackage{dirtytalk}
\usepackage{amsfonts}

\title{Truly Batch Apprenticeship Learning with Deep Successor Features}

\author{
Donghun Lee $^{*1}$
\and
Srivatsan Srinivasan  \footnote{Both the authors contributed equally to this work.} $^1$\and
Finale Doshi-Velez$^1$
\affiliations
$^1$ Harvard University\\
\emails
\{donghunlee,srivatsanstinivasan\}@g.harvard.edu,
finale@seas.harvard.edu
}

\begin{document}

\maketitle
\begin{abstract}
We introduce a novel apprenticeship learning algorithm to learn an expert's underlying reward structure in off-policy model-free \emph{batch} settings. Unlike existing methods that require a dynamics model or additional data acquisition for on-policy evaluation, our algorithm requires only the batch data of observed expert behavior. Such settings are common in real-world tasks---health care, finance or industrial processes ---where accurate simulators do not exist or data acquisition is costly. To address challenges in batch settings, we introduce Deep Successor Feature Networks(DSFN) that estimate feature expectations in an off-policy setting and a  transition-regularized imitation network that produces a near-expert initial policy and an efficient feature representation. Our algorithm achieves superior results in batch settings on both control benchmarks and a vital clinical task of sepsis management in the Intensive Care Unit.
\end{abstract}

\section{Introduction}
Reward design is a key challenge in Reinforcement Learning (RL).  Manually identifying an appropriate reward is often difficult, and poorly specified rewards could lead to serious safety threats \cite{leike2017ai}. Apprenticeship learning is the process of learning how to act from expert demonstrations. To achieve this, Imitation Learning (IL) algorithms - e.g.\cite{ho2016}, directly seek to learn a policy from these demonstrations.  However, directly learning a policy can be brittle in cases of long-horizon planning, \cite{piot2013} environments with strong co-variate shifts and dynamics shifts \cite{fu2017}.  In contrast, Inverse Reinforcement Learning (IRL) approaches \cite{abbeel2004} aim to recover the expert policy by learning the expert's underlying reward function and are often more robust. Explicitly learning the expert's reward function can also inform what the expert wishes to \emph{achieve}, rather than simply what they are \emph{reacting to}, enabling agents to understand and generalize these "intentions" when encountering similar environments.

In this work, we focus on IRL in \textbf{\emph{batch}} settings: we must infer the reward function that the expert had optimized for, given \emph{only} a fixed collection of expert demonstrations. Performing analyses on batch data is desirable and often, the only viable alternative in domains such as health care, finance, education, and industrial automation---situations in which pre-collected logs of expert behavior are relatively plentiful but new data acquisition or a policy roll-out is costly or risky.

While there exist many algorithms for IRL in on-policy settings \cite{abbeel2004,ziebart2008,fu2017}, IRL in batch settings has additional challenges.  Core to many IRL algorithms is the notion of \emph{feature expectation}, or the expected cumulative "feature visits" induced by a policy on a given feature space.  Assuming a well-engineered feature space, the difference between feature expectations from a candidate policy and an expert policy can be used to improve the estimate of the expert reward function\cite{abbeel2004}; In non-batch settings, feature expectations of any proposed candidate policy are computed on transitions collected from its on-policy roll-outs. However, in truly batch settings, we neither have an explicit transition dynamics model nor any ability to acquire new data via on-policy roll-outs. Thus, feature expectations must be estimated off-policy and poor estimates would lead to poor reward updates, rendering IRL ineffective. We also expect batch data to be relatively limited in size and cover a narrow portion of the state-action space and hence, any off-policy estimation algorithms that are sensitive to the distribution of data are expected to generate poor evaluations in truly batch settings. 

Our work makes two key contributions that make truly batch IRL viable.  First, we introduce a model parametrized by a neural network that estimates feature expectations in a completely \emph{off-policy} setting, which we term Deep Successor Feature Network (DSFN). Secondly, we introduce  Transition Regularized Imitation Learning (TRIL) that warm-starts our IRL algorithm with an effective feature representation and a near-expert policy to ensure that candidate policies evaluated by DSFN are not far-off from the expert policy which yielded our batch data. To our knowledge, our work is the first to provide an effective IRL algorithm that scales well across both simple (e.g. control) and complicated (e.g. clinical treatment) environments in completely batch, off-policy, model-free settings.  We demonstrate the effectiveness of our method in benchmark control tasks such as Mountaincar, Cartpole and Acrobot and in a vital clinical problem of managing Sepsis in the Intensive Care Unit.  

\section{Related Work}
Most IRL techniques fall into one of the two categories: margin-based optimization \cite{abbeel2004} or probabilistic optimization \cite{ziebart2008}. In this work, we adopt margin-based optimization, which relies on feature expectations,though all our ideas could be adapted for probabilistic-optimization approaches as well.

\paragraph{Feature Expectations and batch IRL} 
Most IRL works until now have assumed access to a simulator to perform on-policy rollouts \cite{abbeel2004,ziebart2008,fu2017} and relatively few works have considered IRL in a truly batch setting. Like our work, \cite{klein2011} view estimating feature expectations as a policy evaluation problem. Their work proposes Least-Squares Temporal Difference(LSTD) methods and thus inherits the common weaknesses of least squares estimators - a high sensitivity to basis features and the distribution of training data \cite{lagoudakis2003}.  \cite{klein2013} proposed Structured Classification IRL (SCIRL) that optimizes reward by setting action value function as a score metric of a multi-class classification problem. While it is simple in formulation, it still requires estimation of feature expectations done in model-free settings via LSTD methods. Contrary to these LSTD based methods in batch settings, our model uses the representation power of neural networks and prioritized experience replay \cite{schaul2015prioritized} in our DSFN to perform off-policy estimations of feature expectations more effectively.


\paragraph{Warm-Starting IRL with features and Initial Policy} In general, learning a good feature space is instrumental in the success of any IRL algorithm and experts may not always be able to comprehensively specify features characterizing an environment \cite{levine2010}. Attempts to learn rich basis features without manual engineering have been made --- for instance, using hidden layers of neural networks as latent feature encoders \cite{jin2015inverse}. Our model is built along similar lines to use a TRIL network whose hidden layers automatically provides us a feature transformer for our state-action inputs that are fed into the IRL loop. While imitation learning has evolved largely as a non-RL analogue to IRL for learning from expert demonstrations \cite{ross2010efficient,ho2016}, works such as \cite{piot2014boosted} showed the theoretical connections between IRL and IL and proposed a unification framework to help combine advances in these two previously independent domains. Also, other salient IRL works such as \cite{fu2017} have observed the benefits of warm-starting IRL policies with supervised learning. In a similar vein, our TRIL network learns a good initial policy to warm-start IRL --- an indispensable step in batch settings since we have data collected only from the expert policy (Details in Section \ref{sec:methods}).



\section{Background}\label{sec:background}
\paragraph{A. Markov Decision Process :} An MDP is a 5-tuple $(S, A, T, R, \gamma)$ parameterized by (in this work, continuous) states $s \in S$, (discrete) actions $a \in A$, transition probabilities $T(s'|s,a)$, the initial state distribution $d(s_0)$, reward function $R(s,a)$, and discount factor $\gamma \in [0, 1)$. A policy $\pi(a|s)$ is a stochastic map that denotes the probability of taking an action $a$ in state $s$. The value function $V^\pi(s) = \mathbb{E}_\pi[\sum_{t=0}^T \gamma^t r(s_t, a_t) | s_0 = s]$ and the action-value function $Q^\pi(s,a) = R(s,a) + \mathbb{E}_\pi[\sum_{t=1}^T \gamma^t r(s_t, a_t) | s_0 = s, a_0 = a]$ measure the quality of states and actions under any policy $\pi$.  Here, $\mathbb{E}_\pi$ refers to the expectation under the transition dynamics induced by $\pi$ --- $ s_{t+1} \sim T(s_{t+1} | s_t, a_t \sim \pi)$. Finally, $\pi_e$ denotes the expert (optimal) policy such that $\pi_e = \arg\max_\pi V^\pi(s), \forall s \in S$.  

\paragraph{B. Max-Margin IRL and Feature Expectations :}
We assume that we are given $\mathcal{D}=\{(s_0,a_0,...,s_T)\}$, a collection of trajectories sampled according to $\pi_{e}$.  In max-margin IRL \cite{abbeel2004}, we also assume the reward function is linear in some state-action features $R(s,a) = w^\top \cdot \phi(s,a)$ where $\phi(s,a) \in \mathbb{R}^d$ is a feature map defined over $S \times A$. The feature expectations $\mu^\pi(s,a)$ (also known as a successor feature \cite{barreto2017}) for a state action pair under any policy $\pi$ is defined as the expected discounted accumulated ``feature visitations" induced by $\pi$. The overall feature expectation $\mu^\pi$ is defined as the expected $\mu^\pi(s,a)$ over the set of initial states $\mathcal{S}$
\begin{equation}\label{eq:mu_pi}
\begin{split}
    \mu^\pi(s_0,a_0) &=  \phi(s_0,a_0) + \mathbb{E}_\pi \biggl[\sum_{t=1}^{\infty} \gamma^t \phi(s_t,a_t \sim \pi) \biggr] \\
    \mu^\pi &= \mathbb{E}_\mathcal{S} \biggl[\mu^\pi(s_0 \sim \mathcal{S}, a_0 \sim \pi)\biggr]
\end{split}
\end{equation}

 If the reward function is linear in $\phi$, i.e. $R(s,a) = w^T\cdot\phi(s,a)$, the convergence of our agent's feature expectations $\mu^\pi$ to the expert's feature expectations $\mu^\pi_e$ is a sufficient condition for learning a reward structure whose optimal policy matches the expert's policy. \cite{abbeel2004}.

\section{Method: IRL with Deep Successor Features} \label{sec:methods}
While our batch IRL framework is not restricted to one particular IRL algorithm, we adopt max-margin Apprenticeship Learning \cite{abbeel2004} as our IRL algorithm in this work \footnote{Note that even the more recent IRL procedures such as adversarial IRL \cite{fu2017} cannot function without on-policy rollouts to evaluate candidate policies. Future work would involve extending our ideas to more complicated IRL algorithms}. In such max-margin algorithms (Algorithm 1), computing the feature expectations (line 2)  is a key step to evaluate candidate policies. Most max-margin IRL approaches \cite{abbeel2004,ratliff2006} assume an ability to perform on-policy roll-outs(using simulators) or the knowledge of model dynamics to collect additional data---both non-existent in batch settings. In this work, our primary aim is to tackle this inability of performing on-policy rollouts(to evaluate policies) and not to introduce any advancements over IRL algorithms that are already successful in non-batch settings.

Inspired by the linear least-squares approach of \cite{klein2011} to estimate $\mu^\pi$, we interpret the problem of estimating feature expectations in batch settings as an off-policy evaluation problem, drawing a parallel between the feature expectations $\mu^\pi$ (Equation \ref{eq:mu_pi}) as cumulative feature visits and the action value function $Q^\pi(s,a)$(Section \ref{sec:background} A.) as cumulative rewards under a policy $\pi$. This parallel allows us to leverage advances in off-policy action-value function approximation for feature expectation estimation and thus, in Section~\ref{subsec:DSFN}, we introduce Deep Successor Feature Networks (DSFN) as an analogue to Deep-Q networks \cite{mnih2015} in the feature space. 

\begin{figure}
    \centering
    \includegraphics[width=6.75cm,height=4.75cm]{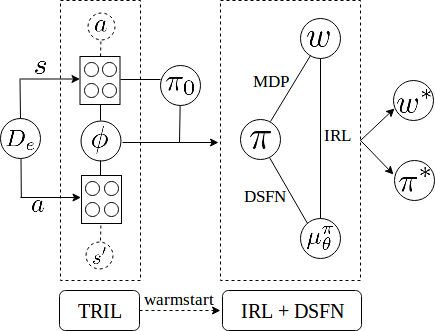}
    \caption{\textbf{Schematic overview of TRIL+DSFN}. TRIL is a dual-channel network that shares certain hidden layers and jointly predicts expert action(a) and state transitions(s'). TRIL warm-starts IRL by providing an initial policy $\pi_0$ and a feature encoder $\phi$ (the joint hidden layers). Using this feature space in IRL, DSFN provides off-policy estimations of feature expectations of any candidate policy, which is essential to update the reward function in max-margin IRL. We used Apprenticeship Learning to optimize the reward function (IRL) and DQN to obtain an optimal policy (MDP).}
    \label{fig:flow}
\end{figure}

\subsection{Estimating Feature Expectations via Deep Successor Feature Network (DSFN)}\label{subsec:DSFN}

Let $D_e = \{ (s_i,a_i,s'_i) \}_{i=1:N_{D_e}}$ denote the batch data sampled using $\pi_e$. Define $s_T$ as the terminal state. Let $\mu_\theta^\pi(s,a)$ denote the feature expectation estimator parameterized by a neural network ($\theta$) for an evaluation policy $\pi$. The aim is to learn $\mu_\theta^\pi(s,a) \approx \mu^\pi(s,a), \forall (s,a)$ and the model is trained using the TD errors from the Bellman equation. Given $\pi,\phi$, we set the Bellman targets $\forall (s, a, s') \in D_e$ in Equation \ref{eq:4}

\begin{eqnarray} \label{eq:4}
   y^\pi_{(s,a,s')} =
    \begin{cases} 
    \phi(s,a) &  \text{if } s' = s_T \\
    \phi(s,a) + \gamma \mathbb{E}_{a'}[ \mu^\pi_\theta(s', a')] &  \text{otherwise}\\
    \end{cases}
\end{eqnarray}
\noindent Notice $y^\pi_{(s,a,s')}$ is specific to $\pi$ and changes with a change in policy. 
We use mean-square error loss to train our deep successor feature network. For a fixed $\pi, \phi$, the loss and its gradient $\forall (s, a, s') \in D_e$ can be calculated as: 
\begin{equation}\label{eqn:gradient}
    \begin{split}
     \mathcal{L}(\theta,\pi) &= \frac{1}{2} \mathbb{E}_{(s,a,s') \sim D_e}\big [\,\, || \mu_\theta^\pi(s,a) - y^\pi_{(s,a,s')}||^2\,\,\big] \\
     &\approx \frac{1}{N_{D_e}} \sum_{i=1}^{N_{D_e}} ||\mu_\theta^\pi(s_i,a_i) - y^\pi_{(s_i,a_i,s_i')}||^2  \\
     \nabla_\theta \mathcal{L}(\theta,\pi) &= \mathbb{E}_{(s,a,s') \sim D_e} \big[\,\, ( \mu_\theta^\pi(s,a) - y^\pi_{(s,a,s')}) \cdot \nabla \mu_\theta^\pi(s,a)\,\,\big]
     \end{split}
\end{equation}

\noindent The training procedure is exactly analogous to that of deep Q-learning \cite{mnih2015} with a subtle difference that DSFN does policy evaluation while DQN does policy optimization. Since we can't collect additional data in batch settings to estimate the performance of DSFN, we carve out a validation dataset and terminate the training when validation loss $\mathcal{L}_{\text{val}}$ converges under a threshold of $\delta > 0$ (Algorithm 2).

\paragraph{Necessity of warm-starting IRL} \label{subsec:necessity_warm_start} Notice that the expectation is taken with respect to transitions from $D_e \sim \pi_e$ in Eqn. (\ref{eqn:gradient}). This implies that in cases of the candidate poilcy $\pi$ being significantly different from $\pi_e$, the batch data support could be nearly disjoint (i.e. $D\sim\pi, D\cap D_e \approx \emptyset$). Since one cannot collect additional transitions in batch settings, our gradient updates for $\mu^\pi$ would be heavily biased. Consequently, IRL with DSFN may fail to converge. Thus, it is crucial to initialize IRL with a near-expert policy so that $\mu^\pi$ can be accurately evaluated on the part of state-action space seen in $D_e$, as opposed to a random policy that most non-batch IRL algorithms typically begin with.

\begin{algorithm}[h]
\textbf{Input}: $\pi_0$ (Initial Policy), $\phi$ (Feature Transformer)\\ 
\textbf{Parameter}: $w \in \mathbb{R}^{d_w}, \theta$\\ 
\textbf{Output}: $w_{(n)}$ 
\begin{algorithmic}[1] 
\FOR{i = 0 : n}
    \STATE Evaluate $\mu_\theta^{\pi_{(i)}}$ using DSFN (Algorithm 2)
    \STATE Compute a reward function $w_{(i)}$ by solving max-margin QP
    \[ w_{(i)} = \min_{w \in \mathbb{R}^{d_w}} ||w||^2  \]
    \[\text{s.t.} \quad w^T \mu^\pi_j \leq w^T \mu^\pi_e + 1, \, \forall j \in \{1, 2, \dots (i-1)\} \]
    \STATE Optimize MDP (any solver) with respect to $r_{w_{(i)}}(s,a) = w_{(i)}^T \phi(s,a)$ to  obtain $\pi_{(i+1)}$.
\ENDFOR
\RETURN $w_{(n)}$
\caption{Batch Max-Margin IRL}
\end{algorithmic}
\end{algorithm}

\begin{algorithm}[h] 
\textbf{Input}: $D_e$ (Data), $\phi$ (Feature Transformer), $\gamma$, $\pi$, $\delta$\\
\textbf{Parameter}: $\theta$\\
\textbf{Output}: $\mu^\pi_{\theta}$
\begin{algorithmic} [1] 
\STATE Feed $D_e$ to Experience Replay Buffer (ERB).
\STATE Initialize $\theta$ for $\mu_{\theta}(\pi,\gamma)$
\WHILE{$\mathcal{L}_{\text{val}} > \delta$}                   
    \STATE Sample a batch $B=\{(s,a,s')\}$ from ERB.
    \STATE Set $\{ y^\pi_{s,a, s'} \}$ for the batch given $\phi, \gamma$ as in Eqn (\ref{eq:4}).
    \STATE Compute a mini-batch gradient of $B$.
    \STATE Update $\theta$ with gradient descent using Eqn (\ref{eqn:gradient}).
\ENDWHILE
\RETURN $\theta$
\caption{Deep Successor Feature Network (DSFN)}
\label{alg:dsfn}
\end{algorithmic}
\end{algorithm}

\subsection{Warm-starting and Feature Learning via Transition-Regularized Imitation Learning}\label{subsec:TRIL}
We propose Transition-Regularized Imitation Learning (TRIL) as a novel batch IL model to obtain a near-expert initial policy while simultaneously deriving a good feature space encoder for the IRL phase. Our TRIL network is a two-channel network jointly trained to predict the expert's action given state and the system's next state transition given state and expert action. Other works have shown that combining dynamics and action prediction is useful in a.) learning a good imitation policy \cite{oh2015action} or b.) creating representations that  reflect the temporal dynamics of the system \cite{song2016}. In our work, we found that TRIL could be leveraged simultaneously for both engineering an effective feature space and a near-expert initial policy for IRL. Knowing that the joint hidden layers capture key information about expert behavior and system dynamics simultaneously, we use those layers as feature encoders to derive corresponding feature representations $\phi$ for input states in IRL. Also, the policy output by TRIL is fed as $\pi_0$ to warm-start Algorithm 1. 

The training procedure of TRIL is similar to that of a multi-channel supervised classifier with regularization. Let $\theta_{\pi_0}$ be the parameters of TRIL and $L_{\text{ce}}$ be the cross entropy loss for predicting expert's action and $L_{\text{mse}}$ be the mean squared error loss on predicting next state given current state and the expert's action assuming we get these samples from demonstration data $D_e$. Let $\lambda$ be the regularization coefficient that controls the strength of the regularization. The network is trained using the following loss: $\forall (s,a,s') \in D_e$
\begin{equation}
\begin{split} 
L(\theta_{\pi_0}) = & L_{\text{ce}}(a, \pi_0(s)) \; + 
 \lambda L_{\text{mse}} (T_{\pi_0}(s, a), s')
\end{split}
\end{equation}

\noindent Figure \ref{fig:flow} presents the full schematic flow of our model that demonstrates the interplay between TRIL and IRL with DSFN. Notice that TRIL learns $\phi(s)$ which can easily be extended to compose $\phi(s,a)$ for a discrete action problem by concatenating one-hot encodings of the actions.

\section{Experimental Procedure}
\paragraph{Training Details}
For our DSFN model, we first trained a TRIL network for warm start.  We used a 70-30 training-validation split and following \cite{duan2016benchmarking}, included a Gaussian output layer that learns the means and standard deviations for the transition prediction --- necessary to learn the uncertainty in our highly stochastic clinical domain experiment (Section \ref{sec:sepsis-results}). Further training details in terms of TRIL, DSFN model architecture and hyperparameters are provided in the appendix (Table 4). The IRL update was computed with the max-margin algorithm \cite{abbeel2004}(Algorithm 1). 

\paragraph{Baselines}
We considered two baselines which, to our knowledge, are the only IRL algorithms that are well-designed to operate in completely batch settings.  The LSTD-$\mu$+LSPI baseline \cite{klein2011} uses Least Squares Temporal Difference (LSTD), a linear model, to approximate estimates of feature expectations ($\mu^\pi)$, and then Least Squares Policy Iteration (LSPI) as the MDP solver. For training the baselines wherever possible, we used the training procedure and model settings provided in the authors' open source implementation \footnote{\label{note1} https://github.com/edouardklein/RL-and-IRL}.  The SCIRL baseline \cite{klein2012} uses estimated feature expectations as a parameterization of the score function of a multi-class classifier (to predict actions). The parameter vector computed this way defines the reward function of the environment and does not require repetitive solving of the RL problem. To make the comparisons fair, all the algorithms compared were initialized with the same initial policy and feature space (using TRIL).

\section{Results: Control Benchmarks} \label{sec:control}
We considered three standard benchmarks: Mountaincar-v0, Cartpole-v0 and Acrobot-v1.\footnote{https://github.com/openai/gym}  In all cases, the optimal policy was first obtained via on-line learning with a DQN \cite{mnih2015}.  This policy was used to generate demonstration data of varying number of episodes (1, 10, 100, and 1000) as training batch data input. Once the data was collected, we no longer accessed the simulator or collected any additional data for the entirety of our process (IL and/or IRL); Thus, the experiments conformed to the truly batch, model-free setting. 

\begin{figure*} 
\centering
\includegraphics[width=0.33\textwidth]{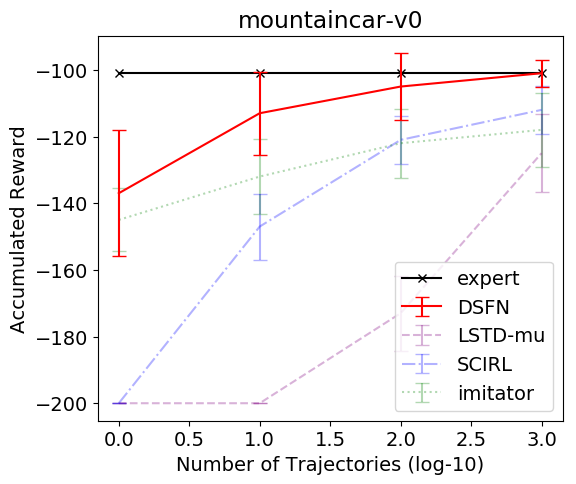}
\includegraphics[width=0.33\textwidth]{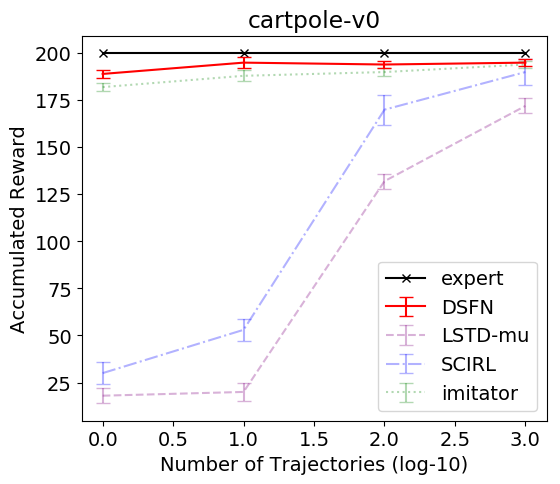}
\includegraphics[width=0.33\textwidth]{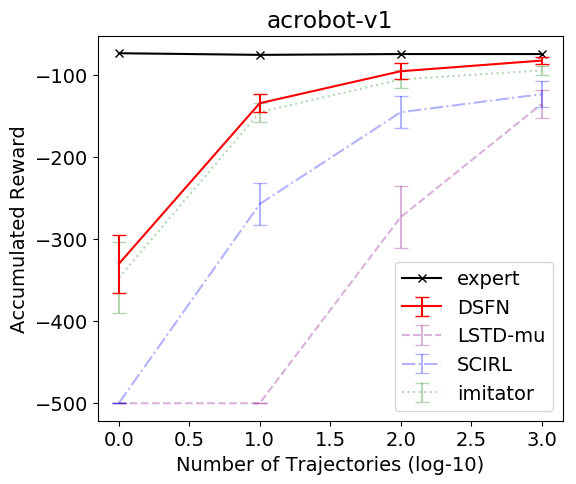}
\caption{Our DSFN learns a reward function that recovers the expert behavior with orders of magnitude less data than the IRL algorithms (note \emph{log} scale on the x-axis).  Its performance is similar or better than the imitator (TRIL without DSFN), which, unlike the IRL approaches, only mimics the policy without recovering the rewards first.  (We emphasize that the IRL approaches have a harder task.)  Performance is averaged over 5 trials where the error bar shows one standard error and as expected, more trajectories lead to less estimation errors.}
\label{fig:control}
\end{figure*}

\paragraph{Our DSFN approach outperformed the baselines with a greater sample-efficiency.} Figure \ref{fig:control} shows the results of our experiments for the three chosen control tasks. Across all tasks and all data regimes, the DSFN model (our model initialized with TRIL) outperforms the baselines, reaching near-expert performance with \emph{an order of magnitude less data}. We observed that $\text{LSTD-}\mu$ performed poorly because of its strong dependence on the coverage and distribution of the input data, which otherwise leads to an under-determined system. We found SCIRL training to be less reliable because it still depended on LSTD methods and shared the same issues and besides, it was hard to fine-tune the hyper-parameters that constitute SCIRL's key heuristic.

\paragraph{Our DSFN approach recovers rewards whose optimal policies match experts similarly or better than imitation learning.}
In figure~\ref{fig:control}, we also compare all the IRL approaches whose goal is to recover the reward function, to a pure imitation learner (imitator-TRIL).  We see that the baselines lag behind the imitation learner, while DSFN matches or exceeds its performance while doing the much harder and useful task of recovering the reward function. While imitation learning is not an IRL approach---and thus not a direct competitor to DSFN, we included this comparison because it answers the key question of whether the features and the feature expectation approximation are expressive and robust enough to find rewards that could recover the expert policy as well as traditional supervised learning. 

\section{Results: Sepsis Management in ICU}\label{sec:sepsis-results}
Sepsis is a leading cause of cost and mortality in Intensive Care Units (ICU), killing 258,000 Americans every year \cite{Singer2016}. Recently, \cite{Raghu2017b} used Deep RL to \emph{optimize} fluid and vasopressor intervention strategies for patients with sepsis. In our work, we focus on learning the rewards associated with choices of vasopressor administration from clinical demonstrations, as vasopressors are a critical clinical intervention to counter the sepsis which often leads to acute hypotension \cite{Singer2016}.

We intend to answer a key question in a complicated problem space: \say{What are clinicians optimizing for with these vasopressor interventions?}  Eliciting a full set of considerations from clinicians is hard, making this an ideal domain for IRL. Understanding their motivations has the potential for building better clinical assistant agents as well as understanding whether ``true'' clinician goals match their stated goals.

\subsection{Problem Set-Up}
\paragraph{Expert demonstrations and MDP definition}
A cohort of 17,898 patients fulfilling Sepsis-3 criteria was obtained from the Multiparameter Intelligent Monitoring in Intensive Care (MIMIC-III v1.4) database \cite{johnson2016mimic}. Our problem setup is similar to the work of \cite{Raghu2017b} which aims to derive optimal policies for sepsis treatment from the available batch data. We model the data comprising 46 features (patient vitals and lab measurements + attributes) including important non-vasopressor interventions such as mechanical ventilation and IV fluids at each time-step as our \emph{continuous state} space i.e. the vector $\textbf{s} \in \mathbb{R}^{46}$. We work in a \emph{discrete action} setting where each action amounted to choosing one among 5 vasopressor dosage bins. We consider in-hospital mortality and leaving the ICU (alive) as the absorbing states (More MDP and feature details can be found in the appendix Sections A.1, A.2).

\subsection{Results}
\paragraph{Imitation: On a real batch IRL task, our DSFN produces approximately 80\% action-matching.}  Our sepsis dataset was divided into a 80-20 train-test partition.  In table~\ref{tab:sepsis_act_match}, we see that the DSFN model achieved significantly higher action-matching rates than the other baselines (setup similar to Section \ref{sec:control}). LSTD-$\mu$, despite heavy tuning, remained sensitive to data distribution and failed to recover good feature expectations as our dataset covered only a narrow portion of the state-action space and SCIRL had similar restrictions because of its dependence on LSTD methods.  We note that all algorithms were given the same features and warm-start (from TRIL) for a fair comparison of imitation results.

\begin{table}
    \centering
    \begin{tabular}{|l|l|l| } 
\hline
Method & Top-1 matching & Top-3 Matching\\
\hline
DSFN & \textbf{$79\pm5\%$} & \textbf{$90\pm3\%$}\\
\hline
LSTD-mu & $39\pm4\%$ & $69\pm3\%$ \\
\hline
SCIRL & $36\pm5\%$ & $61\pm4\%$ \\
\hline
Random & $20\pm1\%$ & $49\pm6\%$ \\
\hline
IL (not regularized) & $29\pm5\%$ & $58\pm4\%$ \\
\hline
\end{tabular}
    \caption{\textbf{Action Matching Probability :} We measured the proportion in which the policy's predicted action fell in the same discrete bin as the ones empirically taken by clinicians. The performance was measured on the test dataset over three trials. Top-1 matching checks whether policy's best action matches clinician actions and the Top-3 matching whether the clinician actions are included in the top 3 choices of the policy.}
    \label{tab:sepsis_act_match}
\end{table}
\begin{figure*}
    \centering
    \includegraphics[scale=0.38]{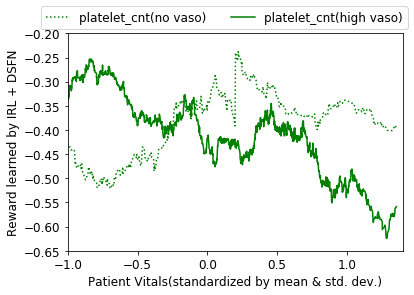}     \includegraphics[scale=0.38]{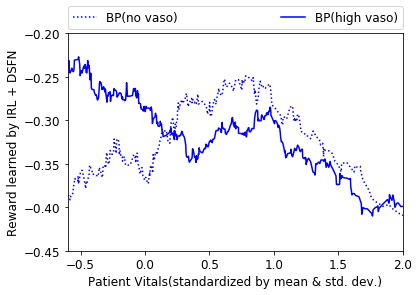}
    \includegraphics[scale=0.38]{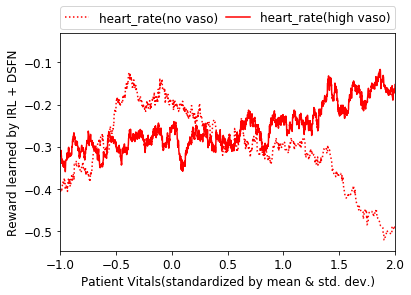}
    \caption{\textbf{IRL with TRIL+DSFN provides clinically intuitive rewards}. Plots of patient vitals(standardized) vs the rewards assigned. On the left, we see that the model places higher rewards on high vasopressor for patients with low platelet counts and no vasopressor is preferred when the platelet counts are stable. In the middle, we see that the model places higher rewards on high vasopressor for patients with low BP and no vasopressor is preferred when BP stabilizes. Similarly, in the right, the model recommends high vasopressor(through rewards) for patients with high heart rate. Remember sepsis shock causes low blood pressure, low platelet counts and high heart rate.}
    \label{fig:vitals_rewards}
\end{figure*}
\paragraph{Interpretability: IRL with DSFN provides insights in line with usual clinical practice.} Action matching is the primary quantitative performance metric to track if we wish to understand whether IRL is finding a reward function consistent with clinical practice. Given that our DSFN model performs relatively well in terms of action-matching, we also focus on the clinical interpretation of the learned rewards in order to verify if the model mimics what the clinicians usually think.  Figure \ref{fig:vitals_rewards} shows the learned rewards with respect to three key patient vitals that are usually extreme  in patients with septic shock. The dashed line indicates the learned reward for doing no action; the solid line for administering a high dose. It is known that patients with sepsis usually suffer from hypotension, high heart rate and low platelet count\footnote{https://www.ncbi.nlm.nih.gov/pmc/articles/PMC1854939/}. We see that our model captures this intuition by penalizing the agent for taking no action when the patients suffer from low BP, high heart rate, or low platelet count and on the contrary, rewards the agent for administering high dosage of vasopressors in such extreme scenarios of septic shock.  These patterns were also verified to be sensible by clinical experts.

\section{Discussion}
IRL in fully batch settings---that is, settings where only a limited, previously-collected set of expert demonstrations are available---is challenging: the data has low coverage over the state-action space, making off-policy estimation tricky, and one may also see variation amongst experts (e.g. in clinical settings).  Our work is one of the first to identify underlying reward functions that recover the expert policy in real, large-scale batch settings (essential, because otherwise we may be interpreting noise) and then interpret them to start understanding \emph{why} experts are making the choices they do.

For both the baselines and our TRIL+DSFN, we found that starting with a good initial policy is crucial for the success of \emph{batch} IRL, especially when the number of expert demonstrations is relatively small.  If the initial proposed policy is drastically different from the expert policy, the IRL algorithms do not converge due to errors propagating from the off-policy feature expectation estimates (details in Section~\ref{subsec:necessity_warm_start}).  However, once initialized in the support of the expert trajectories, our IRL loop converged usually in less than five iterations (details in appendix), and as seen in our results, our model - TRIL+DSFN, is much more robust in its ability to recover the expert reward function from that warm start.  Finally, we note that the warm-start, which produces a policy that closely matches the expert, is \emph{not} IRL : our goal is not to simply recover the expert's policy (a straight-forward supervised problem) but to recover the expert's reward function.  Thus, when DSFN finds a policy similar to the expert policy, it means that it has found a reward function that produces the expert policy under a MDP instead of purely mimicking it.  

While the TRIL+DSFN framework we presented for batch-IRL is very generic, we intend to discuss the contribution of certain key modeling and training choices that enhanced our overall performance. We believe that setting up a transition-based regularization channel jointly with action prediction (TRIL) had certain benefits - a.) learning the dynamics guided the imitation of expert action prediction in line with the system's possible transitions  and b.) the hidden layers that were relevant for both the channels provided a feature transformation that effectively encoded the decisions and temporal dynamics of the problem --- this rich feature space countered the high sensitivity of max-margin IRL to the quality of features \cite{ratliff2006}. Also, since the sepsis environment is highly stochastic, we wanted our DSFN to be aware of the uncertainty estimates for more robust training, which we achieved through the use of Gaussian output layers and we also normalized the states on a rolling basis to provide a consistent range of input values \cite{deepRLmatters2017}.

Broadly, we introduced three key elements that made batch IRL viable - off-policy estimations (DSFN), near-expert initial policy and good feature representations (TRIL) and many IL+IRL algorithms could be fit within this framework. In future, beyond TRIL and DSFN, it would be interesting to explore other methods for identifying feature spaces and warm-starts, as well as other off-policy methods for computing feature expectations---ranging from model-based \cite{herman2016} to importance sampling-based  \cite{thomas2016data}---each of which will have different bias-variance trade-offs.  Finally, we note that our innovations can be combined with other IRL algorithms that use feature expectations, e.g. the entropy-based approaches of \cite{ziebart2008}. 

\section{Conclusion}
We introduced a truly batch IRL method that combines deep successor features, an imitation-based initialization and smart representation learning to effectively recover reward functions that underpin the expert demonstrations. Overall, our model was data-efficient, computation-friendly and comfortably outperformed the baselines with limited demonstrations. Few IRL approaches exist for a truly batch setting, and to our knowledge, ours is the first to work reliably for limited expert demonstrations in large-scale chaotic health-care settings which can be extended to vital problems in other domains such as finance, education and industrial automation.   
 

\begin{thebibliography}{}

\bibitem[\protect\citeauthoryear{Abbeel and Ng}{2004}]{abbeel2004}
Pieter Abbeel and Andrew~Y Ng.
\newblock Apprenticeship learning via inverse reinforcement learning.
\newblock page~1. ACM, 2004.

\bibitem[\protect\citeauthoryear{Barreto \bgroup \em et al.\egroup
  }{2017}]{barreto2017}
Andr{\'e} Barreto, Will Dabney, R{\'e}mi Munos, Jonathan~J Hunt, Tom Schaul,
  Hado~P van Hasselt, and David Silver.
\newblock Successor features for transfer in reinforcement learning.
\newblock In {\em Advances in neural information processing systems}, pages
  4055--4065, 2017.

\bibitem[\protect\citeauthoryear{Duan \bgroup \em et al.\egroup
  }{2016}]{duan2016benchmarking}
Yan Duan, Xi~Chen, Rein Houthooft, John Schulman, and Pieter Abbeel.
\newblock Benchmarking deep reinforcement learning for continuous control.
\newblock In {\em International Conference on Machine Learning}, pages
  1329--1338, 2016.

\bibitem[\protect\citeauthoryear{Fu \bgroup \em et al.\egroup }{2017}]{fu2017}
Justin Fu, Katie Luo, and Sergey Levine.
\newblock Learning robust rewards with adversarial inverse reinforcement
  learning.
\newblock 2017.

\bibitem[\protect\citeauthoryear{Henderson \bgroup \em et al.\egroup
  }{2017}]{deepRLmatters2017}
Peter Henderson, Risashat Islam, Philip BAchman, Joelle Pineau, Doina Precup,
  and David Meger.
\newblock Deep reinforcement learning that matters.
\newblock 2017.

\bibitem[\protect\citeauthoryear{Herman \bgroup \em et al.\egroup
  }{2016}]{herman2016}
Michael Herman, Tobias Gindele, J{\"o}rg Wagner, Felix Schmitt, and Wolfram
  Burgard.
\newblock Inverse reinforcement learning with simultaneous estimation of
  rewards and dynamics.
\newblock In {\em Artificial Intelligence and Statistics}, pages 102--110,
  2016.

\bibitem[\protect\citeauthoryear{Ho and Ermon}{2016}]{ho2016}
Jonathan Ho and Stefano Ermon.
\newblock Generative adversarial imitation learning.
\newblock In {\em Advances in Neural Information Processing Systems}, pages
  4565--4573, 2016.

\bibitem[\protect\citeauthoryear{Jin \bgroup \em et al.\egroup
  }{2015}]{jin2015inverse}
Ming Jin, Andreas Damianou, Pieter Abbeel, and Costas Spanos.
\newblock Inverse reinforcement learning via deep gaussian process.
\newblock 2015.

\bibitem[\protect\citeauthoryear{Johnson \bgroup \em et al.\egroup
  }{2016}]{johnson2016mimic}
Alistair~EW Johnson, Tom~J Pollard, Lu~Shen, H~Lehman Li-wei, Mengling Feng,
  Mohammad Ghassemi, Benjamin Moody, Peter Szolovits, Leo~Anthony Celi, and
  Roger~G Mark.
\newblock Mimic-iii, a freely accessible critical care database.
\newblock volume~3, page 160035. Nature Publishing Group, 2016.

\bibitem[\protect\citeauthoryear{Klein \bgroup \em et al.\egroup
  }{2011}]{klein2011}
Edouard Klein, Matthieu Geist, and Olivier Pietquin.
\newblock In {\em European Workshop on Reinforcement Learning}, pages 285--296.
  Springer, 2011.

\bibitem[\protect\citeauthoryear{Klein \bgroup \em et al.\egroup
  }{2012}]{klein2012}
Edouard Klein, Matthieu Geist, Bilal Piot, and Olivier Pietquin.
\newblock Inverse reinforcement learning through structured classification.
\newblock In {\em Advances in Neural Information Processing Systems}, pages
  1007--1015, 2012.

\bibitem[\protect\citeauthoryear{Klein \bgroup \em et al.\egroup
  }{2013}]{klein2013}
Edouard Klein, Bilal Piot, Matthieu Geist, and Olivier Pietquin.
\newblock A cascaded supervised learning approach to inverse reinforcement
  learning.
\newblock In {\em Joint European Conference on Machine Learning and Knowledge
  Discovery in Databases}, pages 1--16. Springer, 2013.

\bibitem[\protect\citeauthoryear{Lagoudakis and Parr}{2003}]{lagoudakis2003}
Michail~G Lagoudakis and Ronald Parr.
\newblock Least-squares policy iteration.
\newblock volume~4, pages 1107--1149, 2003.

\bibitem[\protect\citeauthoryear{Leike \bgroup \em et al.\egroup
  }{2017}]{leike2017ai}
Jan Leike, Miljan Martic, Victoria Krakovna, Pedro~A Ortega, Tom Everitt,
  Andrew Lefrancq, Laurent Orseau, and Shane Legg.
\newblock Ai safety gridworlds.
\newblock 2017.

\bibitem[\protect\citeauthoryear{Levine \bgroup \em et al.\egroup
  }{2010}]{levine2010}
Sergey Levine, Zoran Popovic, and Vladlen Koltun.
\newblock Feature construction for inverse reinforcement learning.
\newblock In {\em Advances in Neural Information Processing Systems}, pages
  1342--1350, 2010.

\bibitem[\protect\citeauthoryear{Mervyn \bgroup \em et al.\egroup
  }{2016}]{Singer2016}
Singer Mervyn, Deutschman~Clifford S., Seymour Cristopher, and et~al.
\newblock The third international consensus definitions for sepsis and septic
  shock (sepsis-3).
\newblock {\em JAMA}, 315(8):801--810, 2016.

\bibitem[\protect\citeauthoryear{Mnih \bgroup \em et al.\egroup
  }{2015}]{mnih2015}
Volodymyr Mnih, Koray Kavukcuoglu, David Silver, Andrei~A Rusu, Joel Veness,
  Marc~G Bellemare, Alex Graves, Martin Riedmiller, Andreas~K Fidjeland, Georg
  Ostrovski, et~al.
\newblock Human-level control through deep reinforcement learning.
\newblock volume 518, page 529. Nature Publishing Group, 2015.

\bibitem[\protect\citeauthoryear{Oh \bgroup \em et al.\egroup
  }{2015}]{oh2015action}
Junhyuk Oh, Xiaoxiao Guo, Honglak Lee, Richard~L Lewis, and Satinder Singh.
\newblock Action-conditional video prediction using deep networks in atari
  games.
\newblock In {\em Advances in neural information processing systems}, pages
  2863--2871, 2015.

\bibitem[\protect\citeauthoryear{Piot \bgroup \em et al.\egroup
  }{2013}]{piot2013}
Bilal Piot, Matthieu Geist, and Olivier Pietquin.
\newblock Learning from demonstrations: Is it worth estimating a reward
  function?
\newblock In {\em Joint European Conference on Machine Learning and Knowledge
  Discovery in Databases}, pages 17--32. Springer, 2013.

\bibitem[\protect\citeauthoryear{Piot \bgroup \em et al.\egroup
  }{2014}]{piot2014boosted}
Bilal Piot, Matthieu Geist, and Olivier Pietquin.
\newblock Boosted and reward-regularized classification for apprenticeship
  learning.
\newblock In {\em Proceedings of the 2014 international conference on
  Autonomous agents and multi-agent systems}, pages 1249--1256. International
  Foundation for Autonomous Agents and Multiagent Systems, 2014.

\bibitem[\protect\citeauthoryear{Raghu \bgroup \em et al.\egroup
  }{2017}]{Raghu2017b}
Aniruddh Raghu, Matthieu Komorowski, Leo~Celi Ahmed, Imran, Peter Szolovits,
  and Marzyeh Ghassemi.
\newblock Deep reinforcement learning for sepsis treatment.
\newblock 2017.

\bibitem[\protect\citeauthoryear{Ratliff \bgroup \em et al.\egroup
  }{2006}]{ratliff2006}
Nathan~D Ratliff, J~Andrew Bagnell, and Martin~A Zinkevich.
\newblock Maximum margin planning.
\newblock In {\em Proceedings of the 23rd international conference on Machine
  learning}, pages 729--736. ACM, 2006.

\bibitem[\protect\citeauthoryear{Ross and Bagnell}{2010}]{ross2010efficient}
St{\'e}phane Ross and Drew Bagnell.
\newblock Efficient reductions for imitation learning.
\newblock In {\em Proceedings of the thirteenth international conference on
  artificial intelligence and statistics}, pages 661--668, 2010.

\bibitem[\protect\citeauthoryear{Schaul \bgroup \em et al.\egroup
  }{2015}]{schaul2015prioritized}
Tom Schaul, Antonoglou~Ioannis Quan, John, and David Silver.
\newblock Prioritized experience replay.
\newblock 2015.

\bibitem[\protect\citeauthoryear{Song \bgroup \em et al.\egroup
  }{2016}]{song2016}
Zhao Song, Ronald~E Parr, Xuejun Liao, and Lawrence Carin.
\newblock Linear feature encoding for reinforcement learning.
\newblock In {\em Advances in Neural Information Processing Systems}, pages
  4224--4232, 2016.

\bibitem[\protect\citeauthoryear{Thomas and Brunskill}{2016}]{thomas2016data}
Philip Thomas and Emma Brunskill.
\newblock Data-efficient off-policy policy evaluation for reinforcement
  learning.
\newblock In {\em International Conference on Machine Learning}, pages
  2139--2148, 2016.

\bibitem[\protect\citeauthoryear{Ziebart \bgroup \em et al.\egroup
  }{2008}]{ziebart2008}
Brian~D Ziebart, Andrew~L Maas, J~Andrew Bagnell, and Anind~K Dey.
\newblock Maximum entropy inverse reinforcement learning.
\newblock In {\em AAAI}, volume~8, pages 1433--1438. Chicago, IL, USA, 2008.

\end{thebibliography}

 \appendix
 \section{Sepsis}
Here we share the details for the sepsis management experiment. The features that were chosen with a view to represent represent the most important parameters. Clinicians would examine when deciding treatment and dosage for sepsis patients. The features broadly could be categorized into four groups as below.

\subsection{Experimental Details}
When several data points were present in one window, appropriate statistics (mean or sum) deemed apt by clinicians were used for aggregation. The trajectories of clinical measurements have no ``true" state space, so we modeled the data as coming from a continuous state space that consisted of 46 features, including important non-vasopressor interventions such as mechanical ventilation and IV fluids.  We consider in-hospital mortality and leaving the ICU (alive) absorbing states. (Each patient's treatment trajectory comprises an episode of expert demonstrations for our agent to learn from. Our trajectory lengths are less than or equal to 20 steps (about 80 hours of ICU stay since the data was collated over 4 hour bins). Vasopressor actions were discretized into 5 bins: one bin for no dose and 4 associated with quartiles from data.  We used a discount factor $\gamma$ of 0.99.  Our goal was to learn a reward function in this MDP that corresponded to expert behavior. 

\subsection{Patient Features}
\begin{enumerate}
    \item \textbf{Index Measures)} - Shock Index, Elixhauser, SIRS, Gender, Re-admission, GCS - Glasgow Coma Scale, Age
    \item \textbf{Lab Values} - Albumin, Arterial pH, Calcium, Glucose, Hemoglobin, Magnesium, PTT - Partial Thromboplastin Time, Potassium, SGPT - Serum Glutamic-Pyruvic Transaminase, Arterial Blood Gas, BUN - Blood Urea Nitrogen, Chloride, Bicarbonate, INR - International Normalized
    Ratio, Sodium, Arterial Lactate, CO2, Creatinine, Ionised Calcium, PT - Prothrombin Time, Platelets
    Count, SGOT - Serum Glutamic-Oxaloacetic Transaminase, Total bilirubin, White Blood Cell Count
    \item \textbf{Vital Signs}: Diastolic Blood Pressure, Systolic Blood Pressure, Mean Blood Pressure, PaCO2, PaO2, FiO2,  Respiratory Rate, Temperature (Celsius), Weight (kg), Heart Rate, SpO2
    \item \textbf{Intake and Output Events}: Fluid Output - 4 hourly period, Total Fluid Output, Mechanical Ventilation, IV Fluids
\end{enumerate}

\subsection{Discussion on TRIL}
We noticed a significant advantage of having the transition-based regularization (TRIL). As can be seen in Table \ref{tab:sepsis_act_match_app} for sepsis, TRIL outperformed the unregularized baseline. In our sepsis experiment, obtaining an initial policy from TRIL was necessary for DSFN to perform well. DSFN without TRIL did not converge. We think for a task as complex as sepsis management, it is essential to warmstart DSFN with TRIL. We again see the importance of regularization scheme that learns the transition dynamics from its superior performance compared to the unregularized version even though both use the same neural net architecture and training parameters.\\

\noindent For the experiment, we used the same imitation network across all comparisons.  While not an IRL approach, it provides a comparison to how well the agent could do if it did not wish to recover a reward function. We found keeping the policy stochastic to be crucial for this task in line with the multivariate Gaussian scheme described in the main paper. We conjecture this is because sepsis is a complicated disease to manage and even today, there is not a strong agreement  the optimal dosage even within the clinician community and hence learning uncertainty estimates are useful. Another source of prediction errors could be because of the way we discretized our action space, which might not exactly reflect the buckets of vasopressor dosages that clinicians typically operate with while treating patients.

\begin{table}[h]
    \centering
    \begin{tabular}{|l|l|l| } 
\hline
Method & Top-1 matching & Top-3 Matching\\
\hline
TRIL (regularized) & $80\pm2\%$ & $91\pm1\%$ \\
\hline
{IL (not regularized)} & $29\pm5\%$ & $58\pm4\%$ \\
\hline
TRIL + DSFN & \textbf{$79\pm5\%$} & \textbf{$90\pm3\%$}\\
\hline
\end{tabular}
    \caption{\textbf{Sepsis - Action Matching Probability:} We measured the proportion in which the policy's predicted action fell in the same discrete bin as the ones empirically taken by clinicians. The performance was measured on the test dataset over three trials. Top-1 matching checks whether policy's best action matches clinician actions and the Top-3 matching whether the clinician actions are included in the top 3 choices of the policy.}
    \label{tab:sepsis_act_match_app}
\end{table}

\section{OpenAI Control Benchmarks}
Here we share the details for the OpenAI control experiments.

\subsection{Alternate Feature Engineering}
For $\text{LSTD}$-$\mu$ and SCIRL, we also tried other variants of feature engineering by obtaining basis features using the means and standard deviations of the state samples uniformly sampled from the environment. The performance results obtained for the baselines were in the same range as those tabulated in the main paper and hence we do not state the same again.  For MountainCar-v0, we used a Gaussian kernel of 25 components for $\phi(s)$ and subsequently we onehot-encoded $\phi(s)$ based on the 3 actions to represent $\phi(s,a)$ so its dimension becomes 75. For Acrobot-v1 and Cartpole-v0, we used RBF Kernel of 100 components (25 components each $\gamma=0.1, 0.5, 1.0, 5.0$).

\begin{table}[H]
\begin{center}
\begin{tabular}{ |c|c|c| } 
 \hline
 environment & dim(s) & dim(a)\\ 
 \hline
 MountainCar-v0 & 2 & 3\\ 
 Cartpole-v0 & 4 & 2\\ 
 Acrobot-v1 & 6 & 3\\ 
 Sepsis & 46 & 5\\ 
 \hline
\end{tabular}
\caption{\textbf{Benchmark Environments:} state and action space dimensions on OpenAI gym and Sepsis benchmarks}
\end{center}
\label{table:benchmark}
\end{table}

\subsection{Experimental Details}
We set the maximum of 10 iterations with two stopping conditions: first is when feature expectation margin at 0.1 and second is when the difference in validation accuracy for action prediction for the two consecutive iterations drops lower than 5\%. We found the latter stopping condition to be useful in keeping the training loop stable. Unlike typical inverse reinforcement learning routines, there is no correcting mechanism that's based on the ground-truth information (typically achieved by on-policy evaluation) and hence, the training loop may diverge in the complete batch apprenticeship learning.   

\subsection{Neural Network Architectures}
The details can be seen in Table \ref{table:nueral_arch} in the next page. 

\begin{table*}
\begin{center}
\begin{tabular}{ |p{6cm}||p{3cm}|p{3cm}|p{3cm}|  }
  \hline
  Hyperparameters & TRIL & DSFN & DQN\\
 \hline
 number of hidden layers & 2 & 2 & 2\\
 \hline
 hidden node size & 128 & 64 & 128\\
 \hline
 max training iterations & 50000 & 50000 & 30000\\
 \hline
 activation function & tanh & tanh & tanh \\
 \hline
 optimizer & Adam & Adam & Adam \\
 \hline
 adam epsilon & 1e-4 & 1e-4 & 1e-4\\
 \hline
 adam learning rate & 3e-4 & 3e-4 & 3e-4\\
 \hline
 mini-batch size & 64 & 32 & 64\\
 \hline
 $\lambda$ (regularization) & 1.4 & - & -\\
 \hline
 state normalizer & Y & Y & N\\
 \hline
 prioritized experience replay & N & N &  Y\\
 \hline
 prioritized experience replay alpha & - & - &  0.6 \\
 \hline
 prioritized experience replay beta0 & - & - &  0.9 \\
 \hline
 moving average for target network & - & 0.01 & 0.01 \\
 \hline
 discount rate & 0.99 & 0.99 & 0.99\\
 \hline
 stopping condtion (validation) & 5e-3 & 5e-3 & 1e-2\\
 \hline
\end{tabular}
\end{center}
\caption{\textbf{The Hyperparameters of Neural Networks}: to train neural networks, we split the demonstration data into training set (70\%) and validation set (30\%). For the policy network, we found it helpful to establish an isotropic multivariate Gaussian output layer where we output its mean with variable standard deviations for the next state prediction.}
\label{table:nueral_arch}
\end{table*}

\newpage
\newpage
\end{document}